\documentclass{article}

\usepackage[preprint,nonatbib]{neurips_2025_ml4ps}

\usepackage[utf8]{inputenc} 
\usepackage[T1]{fontenc}    
\usepackage{hyperref}       
\usepackage{url}            
\usepackage{booktabs}       
\usepackage{amsfonts}       
\usepackage{textgreek}
\usepackage{siunitx}
\usepackage{microtype}      
\usepackage{xcolor}         
\usepackage{makecell}
\usepackage{physics}
\usepackage{bm}
\usepackage{amsmath}
\usepackage[final]{graphicx}
\usepackage{subcaption}

\title{An Empirical Investigation of Neural ODEs and Symbolic Regression for Dynamical Systems}

\author{Panayiotis Ioannou  \\     %
    Department of Physics \\
    University of Cambridge\\
    \texttt{pi225@cam.ac.uk} \\
    \And
    Pietro Li\`{o} \\
    Department of Computer Science and Technology\\
    University of Cambridge \\
    \texttt{pl219@cam.ac.uk} \\
    \And
    Pietro Cicuta \\
    Department of Physics \\
    University of Cambridge\\
    \texttt{pc245@cam.ac.uk} \\
}

\begin{document}

\maketitle

\begin{abstract}
Accurately modelling the dynamics of complex systems and discovering their governing differential equations are critical tasks for accelerating scientific discovery. Using noisy, synthetic data from two damped oscillatory systems, we explore the extrapolation capabilities of Neural Ordinary Differential Equations (NODEs) and the ability of Symbolic Regression (SR) to recover the underlying equations. Our study yields three key insights. First, we demonstrate that NODEs can extrapolate effectively to new boundary conditions, provided the resulting trajectories share dynamic similarity with the training data. Second, SR successfully recovers the equations from noisy ground-truth data, though its performance is contingent on the correct selection of input variables. Finally, we find that SR recovers two out of the three governing equations, along with a good approximation for the third, when using data generated by a NODE trained on just 10\% of the full simulation. While this last finding highlights an area for future work, our results suggest that using NODEs to enrich limited data and enable symbolic regression to infer physical laws represents a promising new approach for scientific discovery.

\end{abstract} 

\section{Introduction}

Being able to accurately model the dynamics of complex systems or find the differential equations that govern them can help us in understanding those systems better and accelerate scientific discovery. Given the explosion of data and advances in machine learning, a key question arises: how can we best leverage experimental data to infer the dynamics of complex systems? Neural Ordinary Differential Equations (NODEs) \cite{NeuralODE} are a powerful tool for modelling dynamical systems, as their continuous-time nature mirrors that of differential equations. As a result, NODEs have been successfully applied across diverse scientific fields \cite{portwood2019turbulence,lu2021neural}. Much of the existing research has focused on improving NODE architectures \cite{zhang2019anodev2} and assessing their robustness \cite{yan2019robustness}. However, a significant gap remains in evaluating their performance under noisy synthetic data or real-world conditions. Specifically, their interpolation and extrapolation capabilities with respect to boundary conditions have been under-explored. This work addresses this gap by systematically investigating NODE performance using noise-infused synthetic data from two dynamical systems that exhibit damped oscillations: the cart-pole \cite{florian2007correct} and a biological model of bacterial adaptation to changing nutrient environments \cite{droghetti2025incoherent}.

While NODEs excel at modelling dynamics, their black-box nature limits interpretability. In contrast, Symbolic Regression (SR) \cite{cranmer2023interpretable} can discover the exact governing equations; however, it often requires large datasets, which can be a significant challenge in experimental science. To address this, we investigate a pipeline that uses a trained NODE as a data augmentation tool for SR, assessing its ability to recover ground-truth equations from limited initial data. Our evaluation compares the performance of SR on direct ground-truth data (with and without noise) to its performance on a full dataset generated by a NODE trained on only 10\% of the original simulation.

\section{Methods}
This work investigates the extrapolation and interpolation capabilities of NODEs using noisy synthetic data generated from two dynamical systems: the cart-pole \cite{florian2007correct} and a biological system  (which we refer to as the Bio-model) \cite{droghetti2025incoherent}. We also employ SR to determine if the original equations can be recovered directly from the noisy simulation data, as well as from the output generated by the NODE models. The NODE results were obtained using code adapted from the JAX-based \cite{jax2018github} \texttt{Diffrax} library \cite{kidger2021on}. \texttt{PySR} was used for the SR analysis \cite{cranmer2023interpretable}, and the \texttt{phaseportrait} library \cite{phaseportrait} was used to create the phase space plot. All the details to reproduce this work are provided in the remainder of the methods section or in sections \ref{NODETraining}, \ref{PYSRInfo} and \ref{BioExtra} of the appendix.

\noindent\textbf{Cart-pole.}
Equation \ref{AngleCArtpoleEQ} shows the angle dynamics of the cart-pole system assuming no friction between the cart and the rail \cite{florian2007correct}:
\begin{equation} \label{AngleCArtpoleEQ}
    \dv[2]{\theta}{t} = \ddot{\theta} = \frac{g\sin{\theta} + \cos{\theta}(\frac{-F-m_pl\dot{\theta}^{2}\sin{\theta}}{m_c+m_p}) - \frac{\mu_p\dot{\theta}}{m_p l}}{l(\frac{4}{3}- \frac{m_p \cos^{2}{\theta}}{m_c+m_p})}
\end{equation} where $\theta$ is the angle between the vertical and the pole, $\dot{\theta} = \dv[]{\theta}{t}$, $F$ is the force acting on the cart, $l$ is half of the length of the pole, $m_p$ the mass of the pole, $m_c$ the mass of the cart, $\mu_p$ is the friction in the articulation between the cart and the pole and $g$ the gravity. Using a forward simulation, equation \ref{AngleCArtpoleEQ} was solved numerically for many different initial conditions. This was done using the \texttt{SciPy} library, more specifically the \texttt{integrate.odeint} python function. The first data point [$\theta_0$, $\omega_0$] was at $t$ = 0. The simulation ran until $t$ = 10 seconds with $0.01$ second time-steps. Noise of magnitude -5 to 5\% (of the data point value) was added to each point using a uniform distribution. For the results in this report, the simulation was solved using: $F$ = 0, $l$ = 10 cm, $m_p$ = 0.1 kg, $m_c$ = 1 kg and $\mu_p$ = 0.0008. This corresponds to an unforced cart-pole with friction only between the pole and the cart.

Our analysis uses two separate models, each trained on a different set of initial conditions. Model A was trained on 35 boundary conditions, which are all combinations of initial angles from \{0,0.6,1.2,1.8,2.4,3.14\} and angular speeds from \{0,2,4,6,8,10\}, excluding the unstable equilibrium at [0,0]. The training data for Model A consists of 26 points from the first second of simulation for both the angle and angular speed, corresponding to a sampling rate of 25 Hz. The results in Figure \ref{fig:MSEGrid} were generated by Model B, which was trained exclusively on the set of boundary conditions highlighted by the red box in the figure. Its training data was sampled at the same rate and time points as Model A.

\noindent\textbf{Bio-model.}
The system's evolution is governed by the following ordinary differential equations \cite{droghetti2025incoherent}, where each is presented in both a simplified and an expanded form using only the three state variables ($\psi_A$, $\phi_R$, $\chi_R$) and constants ($\phi_{Rmax}, \tilde{\epsilon}, k_\alpha, A, A_2, \tau_\chi$):
{\footnotesize
\begin{gather}
    \dv{\psi_A}{t} = (\phi_{Rmax}-\phi_{R})\nu_f - \lambda[\psi_A,\phi_R](1+\psi_A) = (\phi_{Rmax}-\phi_{R})k_f - \phi_R\tilde{\epsilon}\frac{\psi_A(\psi_A + 1)}{\psi_A + k_\alpha} \label{Biomodel_a_3}\\
    \dv{\phi_R}{t} = \lambda[\psi_A,\phi_R](\chi_R-\phi_R) = \phi_R(\chi_R-\phi_R)\tilde{\epsilon}\frac{\psi_A}{\psi_A + k_\alpha}\label{Biomodel_phi_3} \\
    \dv{\chi_R}{t} = \frac{1}{\tau_\chi}(\omega_R[\psi_A]-\chi_R) = \frac{1}{\tau_\chi}(\frac{\psi_A A}{\psi_A A + A_2}-\chi_R)\label{Biomodel_chi_3}
\end{gather}}
where the simplified form uses $\lambda[\psi_A,\phi_R]$ which is a function of $\psi_A$ and $\phi_R$. The complete model derivation and a comprehensive description of the variables are available in \cite{droghetti2025incoherent}. Appendix \ref{BioExtra} provides a brief overview of the variables and constants. The system's evolution is driven by a shift in nutrient quality, represented by a change in the parameter $\nu$. Each value of $\nu$ corresponds to a unique steady state solution for the state variables ($\psi_A$, $\phi_R$, $\chi_R$). For an example of these values, see Table \ref{table:SSvalues} in the appendix. Experiments begin with the system in a steady state at an initial nutrient quality, $\nu_i$. At $t$ = 0, $\nu$ is abruptly changed to a final value, $\nu_f$, and the system evolves toward a new steady state. This is an up-shift if $\nu_f>\nu_i$ and a down-shift if $\nu_f<\nu_i$. All experiments use a fixed $\nu_f$ = 3.78.

For the numerical simulations, we solved the model for 8 hours with a 0.01-hour time step and added noise to the data in a way similar to the cart-pole. For Model 2A, the training set consisted of 33 data points per hour over the first 4 hours, generated from $\nu_i$ = 2.22 and 3.465. To generate the results in Figure \ref{fig:IntervaltimeExp}, we trained six models, each with a different sampling rate (5, 10, 20, 33, 50, and 100 data points per hour) using data from the first hour of simulation. These models were trained on 12 distinct shifts (6 up-shifts, 6 down-shifts) with $\nu_i$ values ranging from 0.36 to 7.19 in steps of 0.62. The NODE model for SR was trained on the same 12 shifts, but with a sampling frequency of 10 data points per hour from the first 2 hours of each simulation.

\section{Results and Discussion}

\begin{figure}[h]
    \centering
    \begin{subfigure}{0.475\textwidth}
    \centering
    \includegraphics[width=0.90\textwidth]{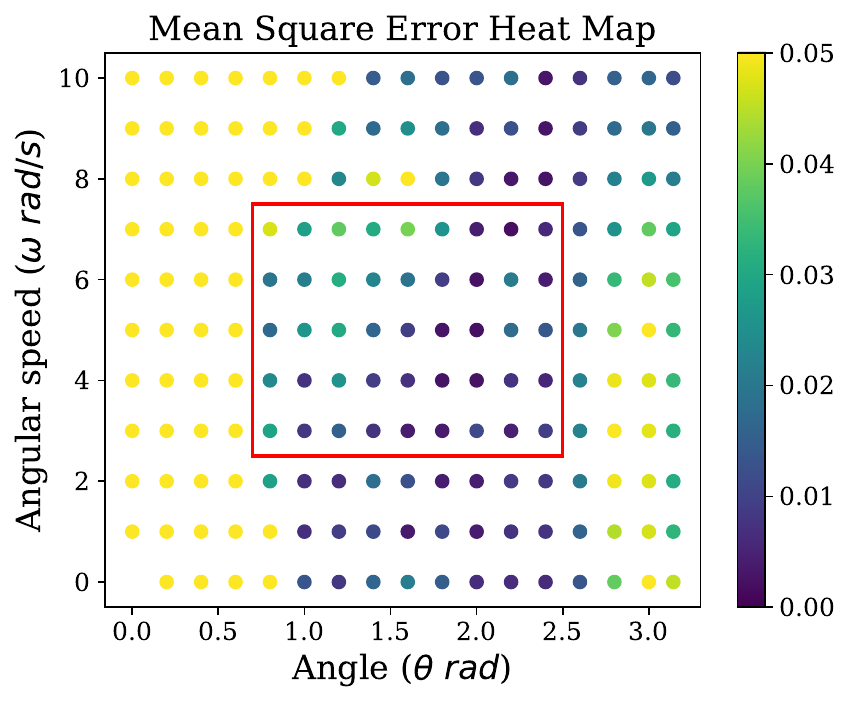} 
    \caption{\label{fig:MSEGrid}}
    \end{subfigure}
    \hfill
    \begin{subfigure}{0.475\textwidth}
    \centering
    \includegraphics[width=0.90\textwidth]{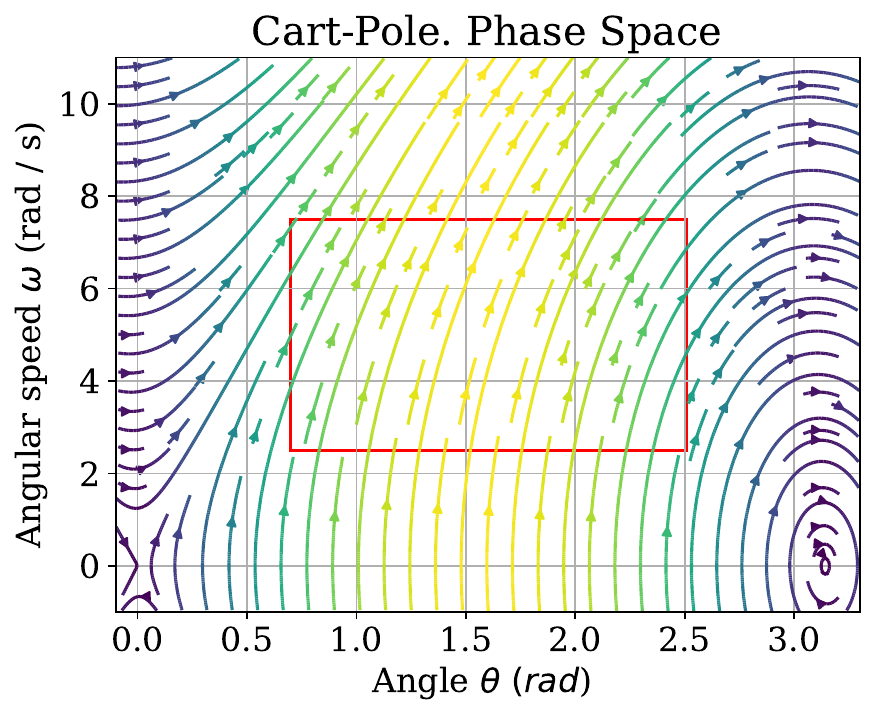}
    \caption{\label{fig:phase_diagram}}
    \end{subfigure}
    \caption{The red rectangle in both plots represents the training region. {\bf (a) The MSE heatmap for Model B (cart-pole).} The low-error zones outside the training region highlight the model's ability to extrapolate. This occurs because the model learned the underlying dynamics from trajectories within the training set, allowing it to accurately predict other points along the same dynamic path. {\bf (b) The cart-pole phase space.} The colour represents the magnitude of the angular speed and acceleration.}
    \vspace{-2mm}
\end{figure}

\begin{figure}[h]
    \centering
    \begin{subfigure}{0.475\textwidth} 
    \centering
    \includegraphics[width=0.90\textwidth]{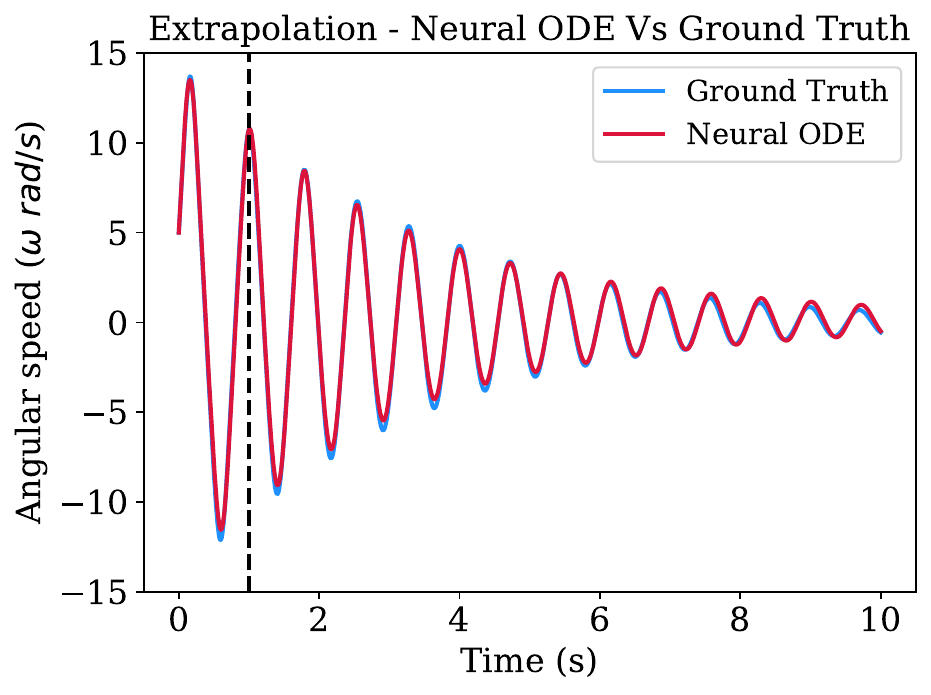}
    \caption{\label{fig:timeextrapolation}}
    \end{subfigure}
    \hfill
    \begin{subfigure}{0.475\textwidth} 
    \centering
    \includegraphics[width=0.90\textwidth]{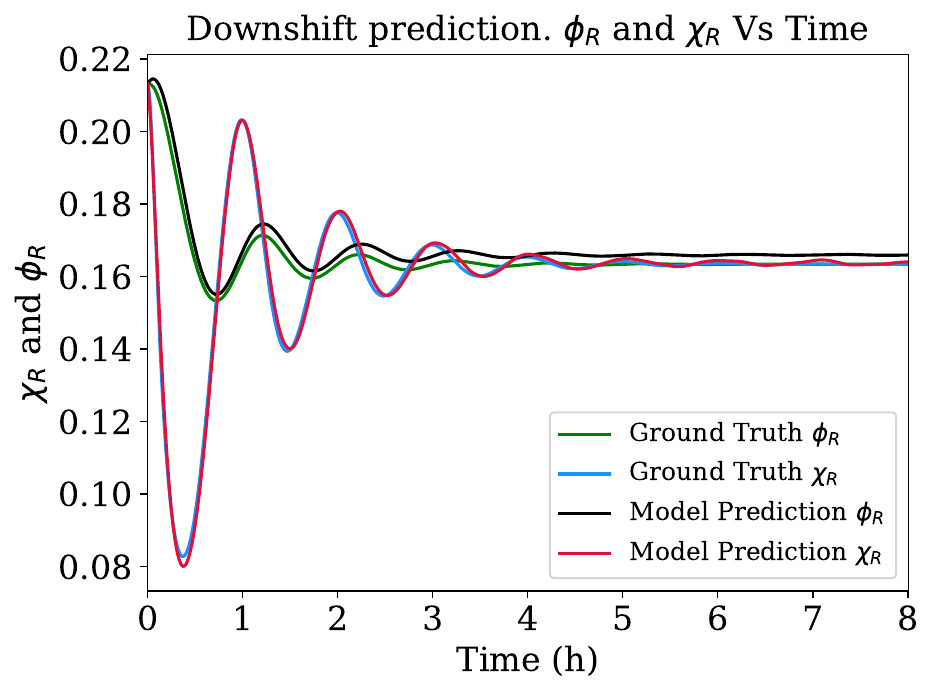}
    \caption{\label{fig:downshift}}
    \end{subfigure}
    
    \caption{{\bf (a) Cart-pole. Model A Time Extrapolation Performance.} Model A, trained on data from only the first second, accurately predicts the system's dynamics for an interpolated initial condition, [1.4, 5]. The model demonstrates strong generalization by successfully extrapolating five times the length of the training data. {\bf (b) Bio-model. Results from Model 2A,} which was trained solely on two up-shift datasets. The model successfully predicts a down-shift ($\nu_i$ = 5.95 to $\nu_f$ = 3.78), an event it was never exposed to in training. While small amplitude errors (less than 5\%) are observed, the model effectively learns and extrapolates the underlying shift dynamics.}
    \vspace{-2mm}
\end{figure}

\noindent\textbf{NODE Extrapolation:}
{\bf(A) Cart-Pole.} Figure \ref{fig:MSEGrid} displays a mean squared error (MSE) heatmap for Model B, where each point represents the initial conditions of a time series. As expected, points within the training region (inside the red rectangle) exhibit a low MSE. Surprisingly, we observe low-MSE regions outside these boundaries, with values comparable to the training set. An analysis of the system's phase space reveals a strong correlation with this phenomenon (Figure \ref{fig:phase_diagram}). Points on the same phase space trajectory as training data points tend to have a lower MSE. This indicates that NODEs can generalise effectively to regions that share the same dynamical properties as the training set. This finding emphasizes a key insight: when designing a training set for dynamical systems, the goal should be to sample diverse dynamics rather than simply a large number of initial conditions. Figure \ref{fig:timeextrapolation} displays the results of model A on an  interpolated, [1.4, 5], initial condition. Trained on data only from the first second, the model accurately captures the dynamics of the system and successfully extrapolates over a duration five times longer than its training period. {\bf(B) Bio-model.} Figure \ref{fig:downshift} displays the performance of Model 2A, which was trained exclusively on only two up-shift simulations. The model accurately predicts the down-shift ($\nu_i$ = 5.95 to $\nu_f$ = 3.78) response, with the error for each point being less than 5\%, despite having no down-shift data included in the training.

\begin{figure*}[h!]
    \centering
    \footnotesize

    \begin{minipage}[h!]{0.497\textwidth}
        
        \centering
        \includegraphics[width=0.80\linewidth]{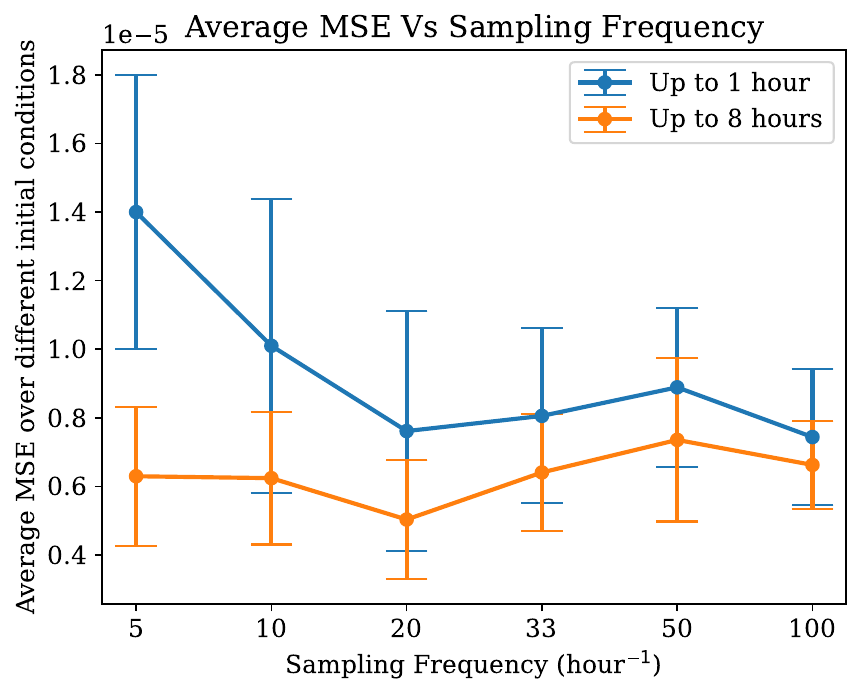}
        \captionof{figure}{{\bf Impact of sampling frequency on model performance.} The legend shows the time regions for MSE calculation. The 8-hour MSE is consistent across all models, demonstrating that long-term predictions are achievable with sparse data. In contrast, the 1-hour MSE rises sharply for the two lowest sampling frequencies, as the model’s fit is highly sensitive to noise when trained on only a handful of points. Errobars indicate the standard deviation over 10 independent runs.}
        \label{fig:IntervaltimeExp}
    \end{minipage}
    \hfill
    \begin{minipage}[h!]{0.497\textwidth}
        \centering
        \vspace{-1mm}
        \captionof{table}{\label{Table} {\bf Results of our SR analysis.} On ground truth data (with $\lambda$ as input), SR recovered all three governing equations but struggled in the presence of 5\% noise, finding only one of the three. On NODE-generated data, SR recovered two of three equations and a good approximation for Equation \eqref{Biomodel_a_3}. Interestingly, it seems that the NODE acts as a denoising filter, enabling better SR performance on the noisy dataset. * The algorithm does not find the correct equation but it finds a decent approximation (see Table \ref{tab:equations}).}
        \footnotesize
        \begin{tabular}{c|c c}
            
            \hline
            \makecell{Equations \\ (with $\lambda$)} & \makecell{Ground Truth \\ No noise} & \makecell{Ground Truth \\ 5\% noise} \\
            \hline
            \ref{Biomodel_a_3} & $\checkmark$ & $\cross$ \\
            \ref{Biomodel_phi_3} & $\checkmark$ & $\checkmark$ \\
            \ref{Biomodel_chi_3} & $\checkmark$ & $\cross$ \\
            \hline
        \end{tabular}
        
        \bigskip
        
        \begin{tabular}{c|c c c}
            \hline
            \makecell{Equations \\ (with $\lambda$)} & \makecell{NODE \\ No noise} & \makecell{NODE \\ 5\% noise} \\
            \hline
            \ref{Biomodel_a_3}  & $\cross$* & $\cross$* \\
            \ref{Biomodel_phi_3}  & $\checkmark$ & $\checkmark$ \\
            \ref{Biomodel_chi_3} &  $\checkmark$ & $\checkmark$ \\
            \hline
        \end{tabular}
    \end{minipage}
\vspace{-4mm}
\end{figure*}

\noindent\textbf{The Impact of Training Data Sampling Frequency on Model Performance.} Figure \ref{fig:IntervaltimeExp} shows the MSE for models trained with varying sampling frequencies on only the first hour of data. The most significant finding is the remarkable consistency of the 8-hour MSE, which shows no significant difference across models. This suggests that high-quality long-term predictions can be achieved with a minimal amount of training data. In contrast, the 1-hour MSE for the two lowest sampling frequencies is notably higher. This is a direct result of the extreme sparsity of the training data; with as few as six training points per variable per shift, noise disproportionately affects the model's fit, leading to a higher interpolation error. The MSE values are averaged over 10 independent runs for each model and across 19 shifts ($\nu_i$ from 0.98 to 6.57 in steps of 0.31).

\noindent\textbf{Symbolic Regression.}
Our SR analysis, presented in Table \ref{Table}, was conducted on two distinct datasets: ground truth simulation data and data generated by a trained NODE model. First, using ground truth data from a single up-shift simulation ($\nu_i$ = 2.53 to $\nu_f$ = 3.78), we successfully recovered all three target equations when $\lambda$ was included as an input variable. However, when the analysis was limited to the three main state variables ($\psi_A$, $\phi_R$, $\chi_R$), only Equation \ref{Biomodel_chi_3} was recovered. One of the primary reasons for this failure is due to the the rational term within $\lambda = 10.04 \cdot \frac{\psi_A\phi_R}{\psi_A + k_a}$ being effectively masked by the data. With $\psi_A$ being, on average, 8 times larger than $k_a$, the denominator $(\psi_A + k_a)$ is dominated by $\psi_A$. This makes the rational term $\frac{\psi_A\phi_R}{\psi_A + k_a}$ approximately equal to $\phi_R$, hiding the true structure and the constant $k_a$ from the SR algorithm. 

The presence of 5\% noise in the ground truth data significantly hindered discovery. As shown in Table \ref{tab:equations}, SR failed to recover the true structure of Equation \eqref{Biomodel_a_3}, finding only the drastic simplification $1.410 - \lambda$. The mean of $\lambda$ over the shift is equal to $1.410$. The target derivatives naturally tend to zero as the system approaches equilibrium, creating a flat fitness landscape; the addition of 5\% noise further masks this already-weak signal, making discovery more difficult. Full \texttt{PySR} outputs for this equation and dataset can be found in Table \ref{tab:GTnoise} in Appendix \ref{pysr_outputs}.

SR applied to data from our NODE (trained on ground truth data with or without noise) recovered two of the three governing equations, \eqref{Biomodel_phi_3} and \eqref{Biomodel_chi_3}. It failed to recover the full form of Equation \eqref{Biomodel_a_3} but nonetheless found good approximations (Table \ref{tab:equations}). This result also demonstrates the NODE's ability to act as a denoiser, enabling SR to achieve a better result on noisy data. The SR solution derived from the NODE trained on noisy data fails to identify the $-\lambda \psi_A$ term. This failure is attributable to a low signal-to-noise ratio: the term's magnitude ($0.01 < \lambda \psi_A < 0.13$, with a mean of \num{0.045}) is negligible compared to the equation's dominant constant, 2.079. Interestingly, \texttt{PySR} compensates for this omitted term by absorbing its mean value into the discovered constant: the algorithm finds 2.006 instead of 2.079. Extended results from some of the SR runs are available in Appendix \ref{pysr_outputs}.

These preliminary findings can be improved, most critically by extending the SR analysis beyond a single-shift simulation to diverse, multi-condition data. Other avenues include improving the NODE model, as its training data was not fully optimised to maximise generalisation in this work, or using more advanced architectures (e.g., Neural CDEs \cite{cdes}). The SR search itself could also be strengthened with physical priors (e.g., unit matching) or alternative frameworks like SINDy \cite{sindy}.

\begin{table}[t]
\centering
\caption{\textbf{SR results for Equation \eqref{Biomodel_a_3}}. Unlike the other two governing equations (which SR recovered successfully from NODE generated data), this equation's discovery is sensitive to the data source. While SR finds the true structure from Ground Truth data, it fails on NODE-generated data by omitting one of the low signal terms, and it fails completely on the noisy Ground Truth data. The mean value of $\lambda \psi_A$ is 0.045. Loss is the MSE between the SR result and the dataset it was trained on.}
\vspace{2mm}
\label{tab:equations}
    \small
    \begin{tabular}{c|c| c}
    \hline
    \multicolumn{2}{c}{\textbf{Ground Truth for Equation (\ref{Biomodel_a_3})}} & $2.079 - 3.78 \times \phi_R - \lambda - \lambda \times \psi_A$\\
    \hline
    \textbf{Dataset} &\textbf{Loss (MSE)} & \textbf{Symbolic Regression Result}  \\
    \hline
    \makecell{Ground Truth (No noise)} & \num{2.09e-8} & $2.076 - 3.77 \times \phi_R - \lambda - \lambda \times \psi_A$ \\
    \makecell{Ground Truth (5\% noise)} & \num{8.12e-3} & $1.410  - \lambda$ \\
    \hline
    \makecell{NODE (No noise)} & \num{1.91e-5} & $2.024 - 3.54 \times \phi_R - \lambda -  \psi_A$ \\
    \makecell{NODE (5\% noise)} & \num{2.16e-4} & $2.006 - 3.67 \times \phi_R - \lambda$ \\
    \hline
    \end{tabular}
\end{table}

\section{Conclusions and Future Work}

This study investigated the generalization properties of Neural Ordinary Differential Equations (NODEs) and their potential to augment data for symbolic regression (SR). Using noisy, synthetic data from two damped oscillatory systems, we demonstrated that NODEs can learn the underlying dynamics and generalize well to novel initial conditions and time periods, provided the trajectories for the new conditions are dynamically similar to those in the training data. Our results, based on sparse and noisy data, highlight that training on diverse dynamics is crucial for generalization, surpassing the need for dense sampling.
We also demonstrated that SR can recover the true governing equations from this noisy data, although its performance is sensitive to the choice of input variables and equation complexity. We tested a pipeline for data-scarce settings: a NODE, trained on only 10\% of the data, generated a full dataset for SR. This pipeline successfully recovered two of the three governing equations and found a good approximation for the third. This pipeline is promising, as these outcomes can be improved with future work, such as implementing techniques like unit matching within the SR framework, using more augmented data or exploring other SR frameworks. Ultimately, this research lays the groundwork for a promising scientific discovery tool, particularly for data-scarce domains, by using NODEs to enrich limited experimental data and enable symbolic regression to infer underlying physical laws.

\begin{ack} 
P.I. was supported by the Engineering and Physical Sciences Research Council Centre for Doctoral Training in Sensor Technologies for a Healthy and Sustainable Future [EP/S023046/1]. We would like to thank Rossana Droghetti and Marco Cosentino Lagomarsino for helpful discussions regarding the Bio-model. Large language models, such as Gemini and ChatGPT, were used to enhance clarity and improve language use.
\end{ack}

\bibliographystyle{ieeetr}
\bibliography{ref}

\newpage

\appendix
\setcounter{table}{0}
\renewcommand{\thetable}{A\arabic{table}}

\section*{Appendix}
The Appendix includes:
(A) details on the training methodology for the Neural ODE models;
(B) a description of the setup and parameters used for the symbolic regression (SR) analysis;
(C) additional details on the Bio-model, including definitions of key variables, equations, and constants; and
(D) the SR outputs from PySR using the data generated from the NODE model that was trained on the ground truth data with 5\% noise and for equation \ref{Biomodel_a_3} also the noisy ground truth data.

\section{Neural ODE Training Details} \label{NODETraining} 
All models were trained using a standard multi-layer perceptron architecture with two hidden layers, each containing 20 nodes. The \texttt{Adabelief} optimizer from the \texttt{optax} library was used for all training runs.

\textbf{Cart-Pole System.} For the cart-pole experiments, all models were trained for 100,000 iterations with a learning rate of 0.003. Model A utilized a batch size of 16, whereas Model B used a batch size of 20.

\textbf{Bio-model.} For the biological system models, Model 2A was trained for 100,000 iterations with a learning rate of 0.001 and a batch size of 1. The Models used to generate the results in Figure \ref{fig:IntervaltimeExp} were trained for 100,000 iterations with a learning rate of 0.003 and a batch size of 10. To generate the data for the SR analysis, the Neural Ordinary Differential Equation (NODE) models were trained for 150,000 iterations with a learning rate of 0.003 and a batch size of 10.  To help prevent convergence to a local minimum, all Bio-model training included an initial 500 iterations using only the first 10\% of the data and a learning rate of 0.003.

\section{Symbolic Regression Details} \label{PYSRInfo}
We employed the \texttt{PySR} library (v1.5.9) to perform symbolic regression and identify the governing equations of the system. The search space for candidate equations included a standard set of primary binary operations (addition, subtraction, multiplication, division) and the unary inverse operator, \texttt{inv(x)=1/x}. We ran the search for 800 iterations with a population size of 50, setting the maximum equation size to 25 and constant complexity to 2.

To serve as the target values for the symbolic regression, the time derivatives of the three state variables ($\psi_A$, $\phi_R$, $\chi_R$) were calculated from the data using the \texttt{NumPy} \texttt{gradient} function. To improve data fidelity, we truncated the time series by removing the first 10 datapoints (0.1 h) and the final 100 datapoints (1.0 h). This preprocessing mitigates two issues: (1) high numerical error from the finite-difference gradient at the initial points, and (2) uninformative, steady-state data, which introduces a low signal-to-noise ratio that can hinder the SR search. While this approach provided a sufficient approximation, using a more robust gradient estimation technique could further improve the accuracy of our symbolic regression results.

\section{Bio-model: Additional information} \label{BioExtra}
A complete description of the model, along with all variables, constants, and the full derivation, is available in \cite{droghetti2025incoherent}. The model, described by equations \ref{Biomodel_a_3}, \ref{Biomodel_phi_3}, and \ref{Biomodel_chi_3} , simulates the adaptive behaviour of \textit{E. coli} bacteria in response to changing nutrient conditions. The system is defined by three key state variables:
\begin{itemize}
    \item $\psi_A$: The ratio of free amino acid mass to the total protein mass within the bacteria.
    \item $\phi_R$: The mass fraction of the proteome dedicated to all ribosome-related proteins.
    \item $\chi_R$: The gene regulatory function, which quantifies the fraction of protein synthesis allocated for ribosome production.
\end{itemize}
The bacterial growth rate, $\lambda$, is a key parameter defined by the relationship $\lambda = \epsilon\phi_R$. Here, $\epsilon$ represents the translation rate, and is given by the function $\epsilon = \tilde{\epsilon} \frac{\psi_A}{\psi_A + k_a}$. The constant $k_a$ = 0.005 is the amino acid uptake efficiency, and  
$\tilde{\epsilon}$ = 10.04 h$^{-1}$ is the theoretical maximum elongation rate. The system's behaviour is governed by the nutrient quality parameter, $\nu_f$, which is solely dependent on the potency of the post-shift nutrient. In this work, we set $\nu_f$ = 3.78 h$^{-1}$. In the steady state, where the time derivatives are zero, each value of $\nu$ corresponds to two solutions for the state variables ($\psi_A$, $\phi_R$, $\chi_R$). One of these solutions yields a negative value for $\psi_A$, while the other is always positive. Since $\psi_A$ represents a physical quantity and must therefore be non-negative, the negative solution is not biologically viable. Consequently, there exists only one unique and physically meaningful steady-state solution for each value of $\nu$, which also corresponds to a unique growth rate, $\lambda$. Examples of these values are presented in Table \ref{table:SSvalues}.

 The RNA Polymerase allocation on ribosomal genes, $\omega_R$, is given by the equation: $\omega_R[\psi_a] = \frac{K_G}{K_G + [G]}$ where $[G]$ is the (p)ppGpp, guanosine tetraphosphate or pentaphosphate, concentration which is given by $[G] = CG_{Ref}(\frac{\psi_a + k_a}{\psi_a}-1)$. In this formulation, we define the shorthand variables $A = K_G$ and $A_2 =  CG_{Ref}* k_a$. For this work, the constants are set to the following values: $C$ = 4.6, $G_{Ref}$ = 101.46 \textmu M, $K_G$ = 14.5 \textmu M, $\phi_{Rmax}$ = 0.55 and $\tau_x$ = 1/6 h. 

Note that the values for the constants $K_G$ and $G_{Ref}$ differ from those presented in the original reference \cite{droghetti2025incoherent}. The referenced paper used values from newer experimental data, such as $K_G$ = 8.07 \textmu M and $G_{Ref}$ = 55.7 \textmu M. Despite these differences, the core results and conclusions of this work remain unchanged.

\begin{table}[h]
\caption{{\bf Steady-state solutions and their corresponding growth rates.} The table presents the unique, biologically viable steady-state values for the state variables ($\psi_A$, $\phi_R$, $\chi_R$) and their associated growth rates, with each solution determined by a specific value of the parameter $\nu$.}

\centering
\begin{tabular}{ c|c c c }
 \hline
 \multicolumn{4}{c}{At Steady State} \\
 \hline
 $\nu$ & \makecell{growth rate \\ ($\lambda$)} & $\psi_a$ & $\phi_R$ and $\chi_R$  \\ 
 \hline
 2.53 & 1.05 & 0.0233  & 0.127  \\
 3.78 & 1.41 & 0.0314  & 0.163  \\
 5.95 & 1.92 & 0.0436  & 0.213  \\
 \hline
\end{tabular}
 
 \label{table:SSvalues}
\end{table}

\section{Symbolic Regression (PySR) outputs}  \label{pysr_outputs}
The following tables detail the \texttt{PySR} discovery results using data from the NODE model that was trained on ground truth data with 5\% noise. Tables \ref{tab:alleq4}, \ref{tab:alleq5}, and \ref{tab:alleq6} show the outputs for equations \eqref{Biomodel_a_3}, \eqref{Biomodel_phi_3}, and \eqref{Biomodel_chi_3}, respectively. Table \ref{tab:GTnoise} shows the \texttt{PySR} discovery results using the ground truth data with the 5\% noise. For clarity, we display equations up to a complexity of 20. The reported \texttt{Loss} is the MSE between a discovered equation and the input dataset to the SR, not the ground truth. An asterisk (*) indicates the equation with the highest score, representing \texttt{PySR}'s optimal trade-off between accuracy and complexity. \textbf{Bold} highlights the equation whose structure is most similar to the true governing equation.

\begin{table}[h!]
\centering
\caption{PySR results for Equation \eqref{Biomodel_a_3} (NODE, 5\% noise data). \textbf{Bold} highlights the structure most similar to the true governing equation; * is PySR's highest score.}
\label{tab:alleq4}
\tiny
\begin{tabular*}{\textwidth}{@{\extracolsep{\fill}} c c c l}
\toprule
\textbf{Complexity} & \textbf{Loss (MSE)} & \textbf{Score} & \textbf{Equation} \\
\midrule
1 & \num{6.296e-03} & 0.000 & $\psi_A$ \\
2 & \num{4.548e-03} & 0.325 & $-0.0053729056$ \\
3 & \num{3.280e-03} & 0.327 & $\phi_R - \chi_R$ \\
4 & \num{1.109e-03} & 1.084* & $1.4028348 - \lambda$ \\
6 & \num{6.885e-04} & 0.238 & $(1.5672859 - \phi_R) - \lambda$ \\
8 & \num{4.006e-04} & 0.271 & $((1.7317353 - \phi_R) - \lambda) - \phi_R$ \\
\textbf{9} & \num{2.160e-04} & \textbf{0.618} & $
\bm{2.0058749 - (\lambda + (\phi_R \times 3.6670127))}$ \\
11 & \num{9.418e-05} & 0.415 & $(2.1844785 - (\lambda + (\phi_R \times 4.556726))) - \psi_A$ \\
13 & \num{6.408e-05} & 0.193 & $1.8036835 - ((((\phi_R \times \phi_R) \times 13.594807) + \lambda) + \psi_A)$ \\
14 & \num{2.118e-05} & 1.107 & $((\phi_R \times -5.776956) - (\lambda + (\chi_R + -2.515695))) \times 0.69243747$ \\
16 & \num{1.523e-05} & 0.165 & $(((-5.5431213 + \chi_R) \times \phi_R) - ((\chi_R + -2.4505582) + \lambda)) \times 0.7387968$ \\
17 & \num{1.449e-05} & 0.050 & $((\phi_R \times -5.272122) - (((\chi_R \times 0.789321) + -2.398444) + \lambda)) \times 0.74833924$ \\
18 & \num{1.252e-05} & 0.146 & $((\phi_R \times (\chi_R + -5.5318727)) - ((\chi_R + -2.4487023) + \lambda)) \times (0.7895749 - \psi_A)$ \\
19 & \num{1.092e-05} & 0.137 & $((\phi_R \times -5.1778607) - (\lambda + (-2.376086 + (\chi_R \times 0.7470051)))) \times (0.810564 - \psi_A)$ \\
... & ...& ... & ... \\
\bottomrule
\end{tabular*}
\end{table}

%
\begin{table}[h!]
\centering
\caption{PySR results for Equation \eqref{Biomodel_phi_3} (NODE, 5\% noise data). \textbf{Bold} highlights the structure most similar to the true governing equation; * is PySR's highest score.}
\label{tab:alleq5}
\tiny
\begin{tabular*}{\textwidth}{@{\extracolsep{\fill}} c c c l}
\toprule
\textbf{Complexity} & \textbf{Loss (MSE)} & \textbf{Score} & \textbf{Equation} \\
\midrule
1 & \num{2.042e-03} & 0.000 & $\psi_A$ \\
3 & \num{1.761e-04} & 1.225 & $\chi_R - \phi_R$ \\
\textbf{5} & \num{3.123e-05} & \textbf{0.865} & $\bm{(\chi_R - \phi_R) \times \lambda}$ \\
6 & \num{2.494e-05} & 0.225 & $(\phi_R - \chi_R) / -0.7338049$ \\
7 & \num{2.412e-05} & 0.034 & $(\lambda - \phi_R) \times (\chi_R - \phi_R)$ \\
8 & \num{9.552e-06} & 0.926 & $(\chi_R - \phi_R) / (0.9726795 - \chi_R)$ \\
9 & \num{9.493e-06} & 0.006 & $(\phi_R \times -1.3376828) + (\chi_R \times 1.3615563)$ \\
10 & \num{7.311e-06} & 0.261 & $(\chi_R \times \psi_A) - ((\phi_R - \chi_R) \times 1.2554784)$ \\
11 & \num{2.109e-06} & 1.243* & $((-1.1330749 - \chi_R) \times (\phi_R - \chi_R)) - -0.0028907147$ \\
13 & \num{1.591e-06} & 0.141 & $(-1.1011528 - \chi_R) \times (\phi_R + ((\psi_A \times -0.07154481) - \chi_R))$ \\
15 & \num{6.273e-07} & 0.465 & $(\phi_R \times -1.1626418) + (\chi_R \times (((\lambda \times \psi_A) \times \lambda) - -1.1181651))$ \\
16 & \num{6.104e-07} & 0.027 & $(\phi_R - (\chi_R \times (\psi_A + 0.98330754))) \times (-0.6877947 - (\lambda \times 0.3857772))$ \\
17 & \num{3.750e-07} & 0.487 & $(\phi_R \times -1.1737428) + (((\psi_A \times (\lambda \times (\lambda - \psi_A))) - -1.1311564) \times \chi_R)$ \\
18 & \num{1.933e-07} & 0.663 & $(\chi_R \times ((((\lambda \times \lambda) + -0.37667426) \times \psi_A) - -1.1579369)) + (\phi_R \times -1.1897844)$ \\
19 & \num{1.800e-07} & 0.071 & $((((\chi_R \times (\lambda \times \lambda)) - \psi_A) \times \psi_A) + (\phi_R \times -1.1864315)) - (-1.1489322 \times \chi_R)$ \\
... & ...& ... & ... \\
\bottomrule
\end{tabular*}
\end{table}

\begin{table}[h!]
\centering
\caption{PySR results for Equation \eqref{Biomodel_chi_3} (NODE, 5\% noise data). \textbf{Bold} highlights the structure most similar to the true governing equation; * is PySR's highest score.}
\label{tab:alleq6}
\tiny
\begin{tabular*}{\textwidth}{@{\extracolsep{\fill}} c c c l}
\toprule
\textbf{Complexity} & \textbf{Loss (MSE)} & \textbf{Score} & \textbf{Equation} \\
\midrule
1 & \num{3.26e-02} & 0.000 & $\psi_A$ \\
3 & \num{3.25e-02} & 0.001 & $\chi_R - \phi_R$ \\
4 & \num{3.13e-02} & 0.036 & $\psi_A + -0.034593984$ \\
5 & \num{1.78e-02} & 0.568 & $(\psi_A / \phi_R) - \chi_R$ \\
7 & \num{4.25e-03} & 0.715 & $(\phi_R \times -21.217772) + 3.486964$ \\
9 & \num{2.91e-03} & 0.190 & $((\psi_A / (\psi_A + \phi_R)) - \chi_R) / \chi_R$ \\
10 & \num{7.57e-04} & 1.345 & $((\psi_A / (\psi_A + \phi_R)) - \chi_R) \times 4.9153194$ \\
\textbf{11} & \num{9.29e-05} & \textbf{2.098*} & $\bm{((\psi_A / (\psi_A + 0.16037777)) - \chi_R) \times 5.927795}$ \\
13 & \num{4.36e-05} & 0.378 & $((\psi_A / (\psi_A + 0.16027561)) - \chi_R) \times (4.554471 + \lambda)$ \\
15 & \num{2.95e-05} & 0.196 & $((\lambda + \lambda) + 3.1795135) \times ((\psi_A / (\psi_A + 0.16015515)) - \chi_R)$ \\
16 & \num{2.93e-05} & 0.005 & $(((\psi_A / (\psi_A + 0.1601735)) - \chi_R) \times 1.8685215) \times (\lambda + 1.7985661)$ \\
17 & \num{2.21e-05} & 0.282 & $(\lambda + ((\lambda - \chi_R) + 3.3858838)) \times ((\psi_A / (\psi_A + 0.16015515)) - \chi_R)$ \\
18 & \num{1.54e-05} & 0.364 & $(((\psi_A / (\psi_A + 0.16009086)) - \chi_R) \times 2.5824206) \times (\lambda + (1.127972 - \chi_R))$ \\
20 & \num{1.51e-05} & 0.010 & $((\psi_A / (\psi_A + 0.16009125)) - \chi_R) \times (((0.9849129 + (\lambda - \chi_R)) \times 2.8481526) - \chi_R)$ \\
... & ...& ... & ... \\
\bottomrule
\end{tabular*}
\end{table}

\begin{table}[t!]
\centering
\caption{PySR results for Equation \eqref{Biomodel_a_3} (Ground Truth, 5\% noise data). \textbf{Bold} highlights the structure most similar to the true governing equation; * is PySR's highest score.}
\label{tab:GTnoise}
\tiny
\begin{tabular*}{\textwidth}{@{\extracolsep{\fill}} c c c l}
\toprule
\textbf{Complexity} & \textbf{Loss} & \textbf{Score} & \textbf{Equation} \\
\midrule
1 & \num{1.185e-02} & 0.000 & $\psi_A$ \\
2 & \num{1.018e-02} & 0.152 & $-0.0044986876$ \\
3 & \num{9.006e-03} & 0.122 & $\phi_R - \chi_R$ \\
4 & \num{7.639e-03} & \textbf{0.165*} & $\bm{1.4092975 - \lambda}$ \\
6 & \num{7.289e-03} & 0.023 & $0.23274253 - (\chi_R \times \lambda)$ \\
7 & \num{6.872e-03} & 0.059 & $((\psi_A / \chi_R) - \chi_R) - \psi_A$ \\
8 & \num{6.736e-03} & 0.020 & $\psi_A + (0.19986038 - (\lambda \times \chi_R))$ \\
9 & \num{5.970e-03} & 0.121 & $(\psi_A - (\chi_R \times (\chi_R + \psi_A))) / \phi_R$ \\
10 & \num{5.926e-03} & 0.007 & $(\psi_A - ((\chi_R \times \chi_R) \times 1.1969346)) / \phi_R$ \\
11 & \num{5.817e-03} & 0.019 & $(\psi_A - (\chi_R \times (\chi_R + (\chi_R \times \chi_R)))) / \phi_R$ \\
12 & \num{5.734e-03} & 0.014 & $(\psi_A - (\lambda \times ((\chi_R \times 0.83034015) \times \chi_R))) / \phi_R$ \\
14 & \num{5.609e-03} & 0.011 & $(\psi_A - (((\chi_R + \psi_A) \times \lambda) / (1.460721 / \chi_R))) / \phi_R$ \\
15 & \num{5.608e-03} & 0.000 & $(((\psi_A - (\chi_R \times (\chi_R \times \lambda))) / \phi_R) \times 0.7639032) - -0.03282497$ \\
16 & \num{5.562e-03} & 0.008 & $(((\chi_R \times ((\lambda \times \lambda) \times (-0.09359126 / \phi_R))) \times \chi_R) + \psi_A) / \phi_R$ \\
17 & \num{5.547e-03} & 0.003 & $((((\psi_A - ((\chi_R \times \lambda) \times \chi_R)) \times 0.1317986) / \phi_R) - -0.0053570317) / \phi_R$ \\
18 & \num{5.519e-03} & 0.005 & $((\psi_A + (\lambda \times ((\lambda \times ((-0.080160506 / \phi_R) \times \chi_R)) \times \chi_R))) / \phi_R) - \psi_A$ \\
19 & \num{5.472e-03} & 0.009 & $(\psi_A + (\chi_R \times (((\lambda \times (\lambda \times -0.027622199)) / \phi_R) + (0.30923742 - \chi_R)))) / \phi_R$ \\
... & ...& ... & ... \\

\bottomrule
\end{tabular*}
\end{table}

\end{document}